\documentclass[letterpaper,10pt,conference]{ieeeconf}

\IEEEoverridecommandlockouts
\usepackage{booktabs}
\usepackage{multirow}
\usepackage{colortbl}
\usepackage{xcolor}
\usepackage{amsmath}
\usepackage{amssymb}
\usepackage{graphicx}
\usepackage{url}
\usepackage{tabularx}
\usepackage{stfloats}
\usepackage[colorlinks=true,urlcolor=blue,linkcolor=black,citecolor=black,bookmarks=false]{hyperref}
\definecolor{rowgray}{gray}{0.92}
\newcommand{\cmark}{\ensuremath{\checkmark}}
\newcommand{\xmark}{\ensuremath{\times}}
\providecommand{\citep}[1]{\cite{#1}}
\newcolumntype{Y}{>{\centering\arraybackslash}X}

\title{\LARGE \bf AgenticNav: Zero-Shot Vision-and-Language Navigation as a Tool-Calling Harness}
\author{Yijian Li$^{1}$ \quad Changze Li$^{1}$ \quad Han Zheng$^{1}$ \quad Jiyuan Cai$^{2}$ \quad Tong Qin$^{1,*}$ \quad Ming Yang$^{1}$\\
{\normalsize $^{1}$Shanghai Jiao Tong University \quad $^{2}$Yinwang Intelligent Technology Co., Ltd.}}
\hypersetup{
  pdftitle={AgenticNav: Zero-Shot Vision-and-Language Navigation as a Tool-Calling Harness},
  pdfauthor={Yijian Li, Changze Li, Han Zheng, Jiyuan Cai, Tong Qin, Ming Yang}
}
\begin{document}
\maketitle
\begingroup
\renewcommand{\thefootnote}{\fnsymbol{footnote}}
\footnotetext[1]{Corresponding author.}
\endgroup
\thispagestyle{empty}
\pagestyle{empty}

\begin{abstract}
Zero-shot vision-and-language navigation in continuous environments (VLN-CE) has recently become feasible with large vision-language models (VLMs). Existing methods typically rely on learned waypoint predictors to propose navigable actions. This limits the model's action space and fails to leverage depth inputs effectively. Moreover, memory is commonly handled by accumulating long textual or visual histories, which overwhelms the prompt context. In this paper, we rethink zero-shot VLN-CE as an agentic interface between the VLM and the environment, and present \textbf{AgenticNav}, a lightweight navigation harness that exposes action, depth, and memory as callable tools. Instead of choosing from predicted waypoints, the action tool allows the VLM to directly select a target pixel in RGB observations, converting it into executable motion. Depth is exposed through an on-demand pixel-depth tool, enabling the VLM to request precise metric distances only where they matter. For memory, AgenticNav uses an agentic memory architecture combining reasoning text history and a compact map image, paired with a recall tool that allows the VLM to selectively revisit past visual observations. On R2R-CE, AgenticNav achieves state-of-the-art (SOTA) zero-shot performance under same VLM comparison, reaching 76\% success rate and 66.50\% SPL. Ablation experiments validate the effectiveness of our tool and memory designs, while backbone evaluations demonstrate more consistent navigation gains as VLMs improve, highlighting the harness's potential to benefit from future VLM advances. Real-world experiments on two distinct robotic platforms further demonstrate its zero-shot generalization and adaptability.

\end{abstract}

\begin{figure}[!t]
\centering
\includegraphics[width=\columnwidth]{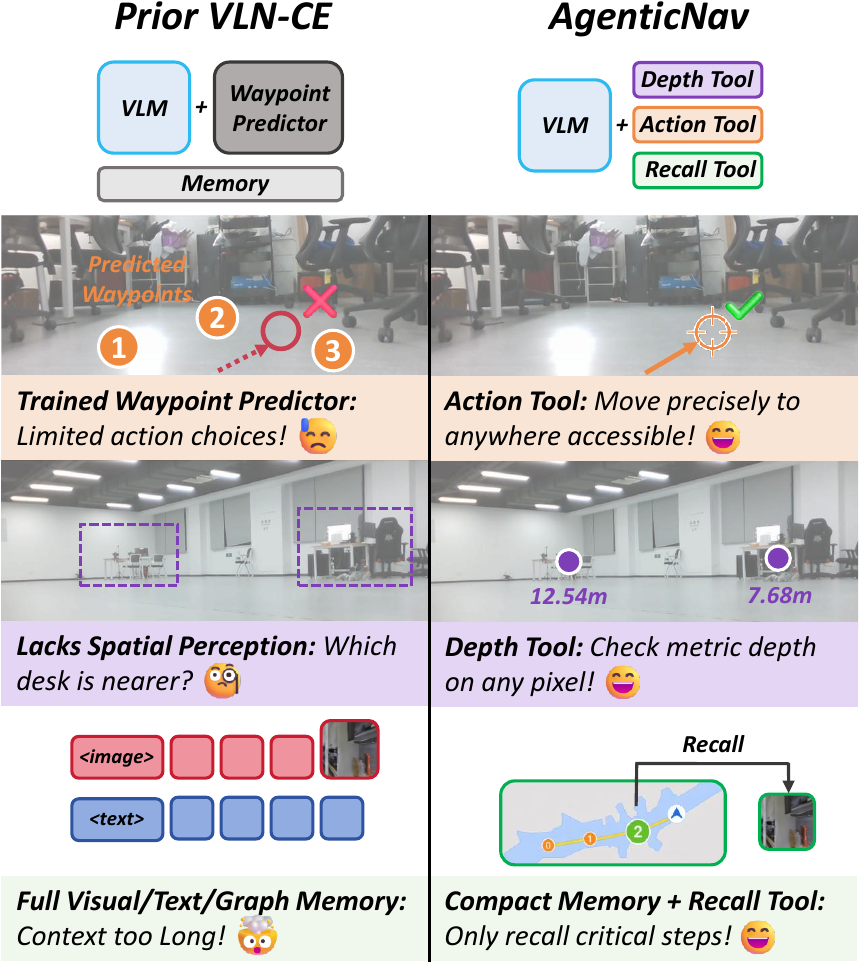}
\caption{Overview of AgenticNav. We reformulate zero-shot VLN-CE as a tool-calling interface that exposes depth, memory, and action grounding to the VLM.}
\label{fig:intro}
\end{figure}


\section{Introduction}

Imagine a robot asked to ``turn right, pass the kitchen, and stop near the sofa'' in an unseen home. Zero-shot vision-and-language navigation in continuous environments (VLN-CE) captures this setting by asking agents to follow natural-language instructions in unseen 3D scenes without relying on predefined navigation graphs~\citep{anderson2018r2r,krantz2020vlnce}. With the rapid development of large vision-language models (VLMs), recent zero-shot VLN-CE systems employ VLM as the high-level decision maker. The core challenge therefore becomes an interface problem: a VLM can reason over language and images, but the robot must eventually interact with the spatially complex environment. 

Existing state-of-the-art methods, including Open-Nav~\citep{qiao2025opennav}, SmartWay~\citep{shi2025smartway}, and EvoNav~\citep{dai2026evonav}, commonly build this interface around a learned waypoint predictor. Such a predictor is an additionally trained visual neural network that takes panoramic RGB and depth observations as input and proposes a small set of nearby navigable points for the VLM to choose from. This design is convenient because it shields the VLM from continuous control. However, it also introduces several bottlenecks that can prevent the VLM from fully using its visual reasoning capability.

First, waypoint predictors restrict the action space. The VLM can only choose among a small set of predicted candidates, so an instruction-relevant location may be unreachable if the predictor does not expose it. Second, waypoint-based interfaces take depth inputs as ground truth but do not explicitly expose them to the VLM to provide effective spatial information. This makes it difficult for the VLM to acquire accurate depth information of specific locations for spatial reasoning.

Memory creates a third bottleneck. Many methods~\citep{qiao2025opennav, shi2025smartway} maintain memory by adding more textual or visual context to the prompt. This can help with progress tracking, but most past context is irrelevant to the current decision and may distract the model as the context grows. Other methods, such as EvoNav~\citep{dai2026evonav}, retrieve cross-episode experience as memory. While useful, this weakens the zero-shot assumption and is not reliable when deployed in novel scenes where no prior episode experience is available.

In this work, we revisit VLN-CE from an agentic tool-calling perspective and introduce \textbf{AgenticNav}, a lightweight harness for strictly zero-shot VLN-CE. Rather than feeding the VLM a fixed set of waypoint candidates, dense depth inputs, or accumulated history, AgenticNav exposes action, depth, and memory as callable tools.

The \textbf{Action Tool} enables waypoint-free action selection. Instead of selecting from predicted waypoints, the VLM directly chooses a target pixel in the RGB image. The tool then converts that selection into an executable action through depth unprojection and safety checking. This keeps numerical geometry outside the VLM while preserving its freedom to select any visible location. The \textbf{Depth Tool} provides on-demand pixel-depth queries. The VLM may query metric depth at selected image locations before acting, enabling targeted comparisons such as whether a corridor is open, whether a doorway is nearer than an obstacle, or which candidate point is reachable. This exposes depth to the VLM as precise local evidence instead of hiding it inside a waypoint predictor or forcing the model to parse a full depth map.

Beyond depth and action tools, AgenticNav organizes memory through reasoning text history, a compact map image, and a selective visual \textbf{Recall Tool}. The reasoning text history preserves past decision rationales, while the map image summarizes the navigation trajectory and indexes historical observations. When the VLM decides that visual details from a past node become useful, it can call the tool to retrieve the corresponding RGB observation. Together, these components preserve key context while providing access to past visual observations on demand.

We evaluate AgenticNav on the R2R-CE \texttt{val\_unseen} 100-episode protocol used by recent zero-shot VLN-CE work. With Gemini-3.7-Flash as the VLM core, AgenticNav achieves 76\% SR and 66.50\% SPL, outperforming the next-best zero-shot method by 22\% SR and 18.55\% SPL. With other VLMs, AgenticNav also demonstrates leading performance under the same benchmark setting. Component ablations further confirm the contribution of our tool interface and memory architecture to navigation performance. Tests on two different robotic platforms also demonstrate that AgenticNav supports real-world navigation across hardware configurations in a zero-shot setting. These results suggest that the limiting factor in zero-shot VLN-CE is no longer only the reasoning model itself, but also the action, depth, and memory harness through which the model acts. In multi-backbone comparisons, AgenticNav exhibits sustained navigation gains as VLM backbones evolve, highlighting a key property of an effective harness: translating advances in its VLM core into better embodied performance.

Our main contributions are:
\begin{itemize}
\item We present \textbf{AgenticNav}, a navigation agent harness that rethinks zero-shot VLN-CE as an agentic tool-calling process. By combining on-demand metric depth queries through the Depth Tool with direct pixel-target selection through the Action Tool, the harness enables the VLM to gather spatial evidence and choose where to move without relying on trained waypoint predictors.
\item We design an agentic memory architecture combining reasoning text history, a compact map image, and a selective visual Recall Tool. It preserves key context and retrieves past observations on demand, avoiding the accumulation of irrelevant historical frames in the VLM context.
\item We demonstrate state-of-the-art (SOTA) zero-shot performance on the R2R-CE benchmark under fair VLM backbone comparisons, surpassing the next-best zero-shot method by 22\% SR and 18.55\% SPL. Ablation experiments validate the effectiveness of our tool and memory designs. Real-world validation further highlights the zero-shot generalization of our method.
\end{itemize}

\section{Related Works}

\begin{figure*}[t]
\centering
\includegraphics[width=0.94\linewidth]{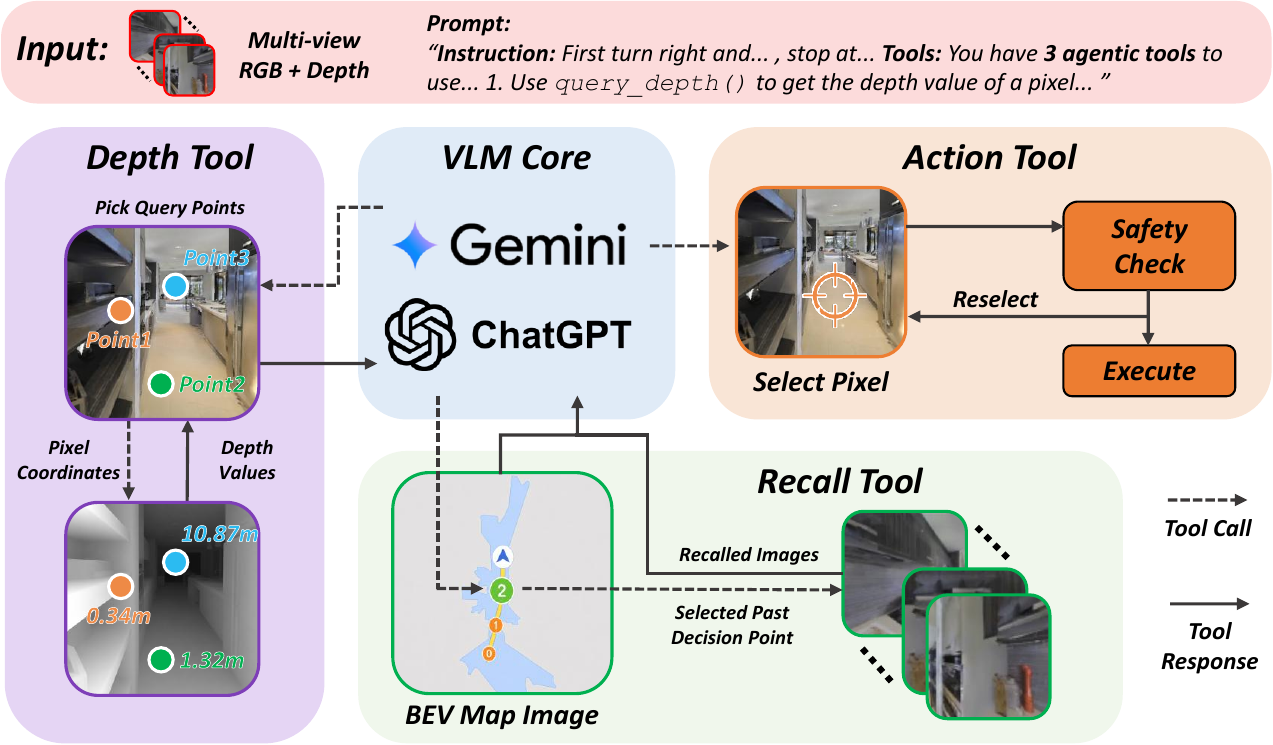}
\caption{AgenticNav method overview. The VLM queries depth, optionally recalls past node images, and commits to an action; safety checking triggers either Reselect or Execute.}
\label{fig:method}
\end{figure*}

\subsection{Vision-and-Language Navigation}
Vision-and-language navigation (VLN) asks an agent to follow natural-language instructions in unseen scenes~\citep{anderson2018r2r}. Early VLN methods mainly follow a training-based paradigm in discrete environments, selecting among predefined graph nodes rather than executing continuous motion~\citep{fried2018speaker,ma2019regretful}. VLN in continuous environments (VLN-CE) removes the graph and requires low-level movement in continuous 3D environments~\citep{krantz2020vlnce}; supervised methods therefore introduce spatial intermediates, such as waypoint predictors, to connect language decisions with executable motion~\citep{krantz2021waypoint,hong2022bridging,chen2022duet,an2023bevbert,an2023etpnav,wang2023gridmm}. Recent end-to-end or large-model approaches, including NaviLLM, Navid, NaVILA, and LightNav-0, scale embodied navigation data or adapt VLM and VLA models for action prediction~\citep{zheng2024generalist,zhang2024navid,cheng2024navila,wang2026lightnav}. Although effective, their dependence on domain-specific training data and action supervision limits cross-domain and sim-to-real generalization, motivating zero-shot navigation with foundation models.

\subsection{LLMs and VLMs for Zero-Shot Navigation}
With the emergence of large language models (LLMs) and VLMs, recent works have explored zero-shot navigation without training a full policy~\citep{ichter2023saycan,song2022llmplanner,chen2023a2nav}. Some approaches~\citep{zhou2024navgpt,long2024discussnav,chen2024mapgpt,long2024instructnav,zhan2024mcgpt} use prompted foundation models for instruction reasoning, mapping, and high-level decision-making across discrete VLN, continuous VLN-CE, and generic instruction-navigation settings. To operate in continuous VLN-CE, methods such as Open-Nav~\citep{qiao2025opennav} and SmartWay~\citep{shi2025smartway} usually rely on trained waypoint predictors, while EvoNav~\citep{dai2026evonav} keeps this design and retrieves experience from previous episodes. These interfaces can restrict action choices to predefined candidates, and growing prompt histories may introduce redundant context. In contrast, AgenticNav addresses these limitations while restricting its memory to information acquired within the current episode.

\subsection{Embodied Agent Harness}
Recent robotics work shows that foundation-model performance depends strongly on the harness that grounds reasoning into perception, memory, actions, and feedback. Some works~\citep{ichter2023saycan,huang2023inner,liang2023code,shah2023lmnav,rajvanshi2024saynav} use LLMs or VLMs as high-level planners over robot skills, affordance functions, executable APIs, maps, or classical controllers, enabling closed-loop grounding without training a full end-to-end policy. Other work studies embodied multimodal models that connect sensor observations and action prediction more directly, but usually require large-scale robot data or task-specific adaptation~\citep{driess2023palme}. ReasonNav~\citep{chandaka2025reasonnav} illustrates the value of an agentic VLM harness with landmark maps and higher-order navigation skills, but targets semantic building navigation rather than instruction-following VLN-CE. AgenticNav instead formulates VLN-CE as explicit tool calling: the VLM selects pixel-level targets, queries metric depth only where needed, and invokes selective visual recall over episodic memory. This gives the VLM a direct, grounded, and continuous interface while preserving zero-shot operation.

\section{Method}
\label{sec:method}

\subsection{Agentic Tool-Calling Interface}
Given instruction $L$, the harness receives a heading-indexed RGB-D multi-view observation $O_t=\{(I_t^k,D_t^k)\}_{k=0}^{N-1}$, where $I_t^k$ is the RGB image and $D_t^k$ is the aligned depth map retained by the tools. In AgenticNav, the VLM is not directly given depth image inputs. At the beginning of decision step $t$, the VLM receives:
\begin{equation}
X_t=(L,\mathcal{I}_t,B_t,H_t).
\end{equation}
Here $\mathcal{I}_t=\{I_t^k\}_{k=0}^{N-1}$ contains the current RGB views, $B_t$ is the compact BEV map image, and $H_t$ contains the reasoning text and action records from recent steps. 

As illustrated in Fig.~\ref{fig:method}, instead of asking the VLM to output low-level controls or choose learned waypoints, AgenticNav exposes four tools:
\begin{equation}
\mathcal{T}=\{\texttt{query\_depth},\texttt{recall},\texttt{move\_to},\texttt{stop}\}.
\end{equation}
The first three tools correspond to the \textbf{Depth Tool}, \textbf{Recall Tool}, and \textbf{Action Tool}, respectively.
At each decision step, the \textbf{VLM Core} may issue multiple tool calls. Calls to \texttt{query\_depth} and \texttt{recall} return additional evidence to the next VLM turn, while successful calls to \texttt{move\_to} and \texttt{stop} are terminal for the current step. Fig.~\ref{fig:context_management} summarizes how these per-step inputs, outputs, and tool responses are organized as context.

\subsection{Depth Tool}
Dense depth images are hard for VLMs to parse quantitatively, so AgenticNav exposes depth through the tool \texttt{query\_depth}$(\mathcal{P})$. The VLM queries a batch of candidate pixels
\begin{equation}
\mathcal{P}=\{p_i=(k_i,u_i,v_i)\}_{i=1}^{n},
\end{equation}
where $n$ is the number of queried pixels, $k_i$ indexes one of the panoramic RGB views, and $(u_i,v_i)\in[0,1]^2$ are normalized image coordinates, with $u_i$ increasing from left to right and $v_i$ increasing from top to bottom. For each queried pixel, the tool samples the aligned depth map and returns
\begin{equation}
\mathcal{D}(p_i)=
\left(d_i,\; s_i^{5 \times 5}\right).
\end{equation}
Here $d_i$ is the metric depth at the selected pixel in metres. The term $s_i^{5 \times 5}$ denotes local depth statistics from a $5 \times 5$ neighbourhood around the pixel, including minimum, mean, and median depth, which help the VLM judge whether the point lies on a stable surface or near a depth boundary. The depth query itself does not move the robot; it only provides numeric evidence for comparing candidate floor, doorway, or landmark-adjacent pixels without requiring the VLM to parse a full depth image.

\subsection{Action Tool}
The Action Tool \texttt{move\_to}$(k,u,v,\texttt{reason})$ selects a normalized pixel target $(u,v)$ in view $k$. The \texttt{reason} field requires the VLM to explain the target selection and is retained with the executed action as a memory record.

Let $(x,y)$ be the image-pixel coordinates corresponding to $(u,v)$, $d=D_t^k(x,y)$ the depth, $K$ the camera intrinsics, and $T_k$ the transform to the agent frame. The target is back-projected as:
\begin{equation}
p_t=T_k\!\left(dK^{-1}[x,y,1]^\top\right).
\end{equation}
Its ground-plane bearing and executable distance are:
\begin{equation}
\begin{gathered}
\theta=\mathrm{bearing}(p_{t,xz}),\\
\rho=\min\!\left(\rho_{\max},\max(0,\|p_{t,xz}\|-m)\right).
\end{gathered}
\end{equation}
Here $m$ is the stopping margin and $\rho_{\max}$ is the maximum distance per action. The target distance is reduced by $m$ and clipped to $[0,\rho_{\max}]$.

The tool rejects out-of-range targets and invalid depth. In the agent frame, $y$ is vertical and $xz$ is the ground plane. Back-projected points within the robot's body-height range form $\mathcal{P}_t=\{q_j\mid y_{\min}\le q_{j,y}\le y_{\max}\}$, whose ground-plane projections give the obstacle set $\mathcal{O}_t=\{q_{xz}\mid q\in\mathcal{P}_t\}$. For forward direction $b_\theta=(-\sin\theta,\cos\theta)$, lateral direction $\ell_\theta=(\cos\theta,\sin\theta)$, and agent radius $r_a$, define the rectangular region checked for obstacles as
\[
\mathcal{R}(\theta,\rho)=
\left\{s b_\theta+h\ell_\theta\;\middle|\;
\begin{gathered}
0<s<\rho+r_a,\\
|h|<r_a
\end{gathered}\right\}.
\]
Here $s$ is the forward distance and $h$ is the lateral offset. The safety check passes when this region contains no observed obstacle points:
\begin{equation}
\mathrm{Safe}(\theta,\rho)=
\mathbf{1}\!\left[\mathcal{O}_t\cap\mathcal{R}(\theta,\rho)=\varnothing\right].
\end{equation}
Rejected motions return \textbf{Reselect} feedback; accepted motions \textbf{Execute} $(\theta,\rho)$, as shown in Fig.~\ref{fig:method}. The \texttt{stop(reason)} tool terminates the episode.

\subsection{Agentic Memory and Recall Tool}
\begin{figure}[!t]
\centering
\includegraphics[width=\columnwidth]{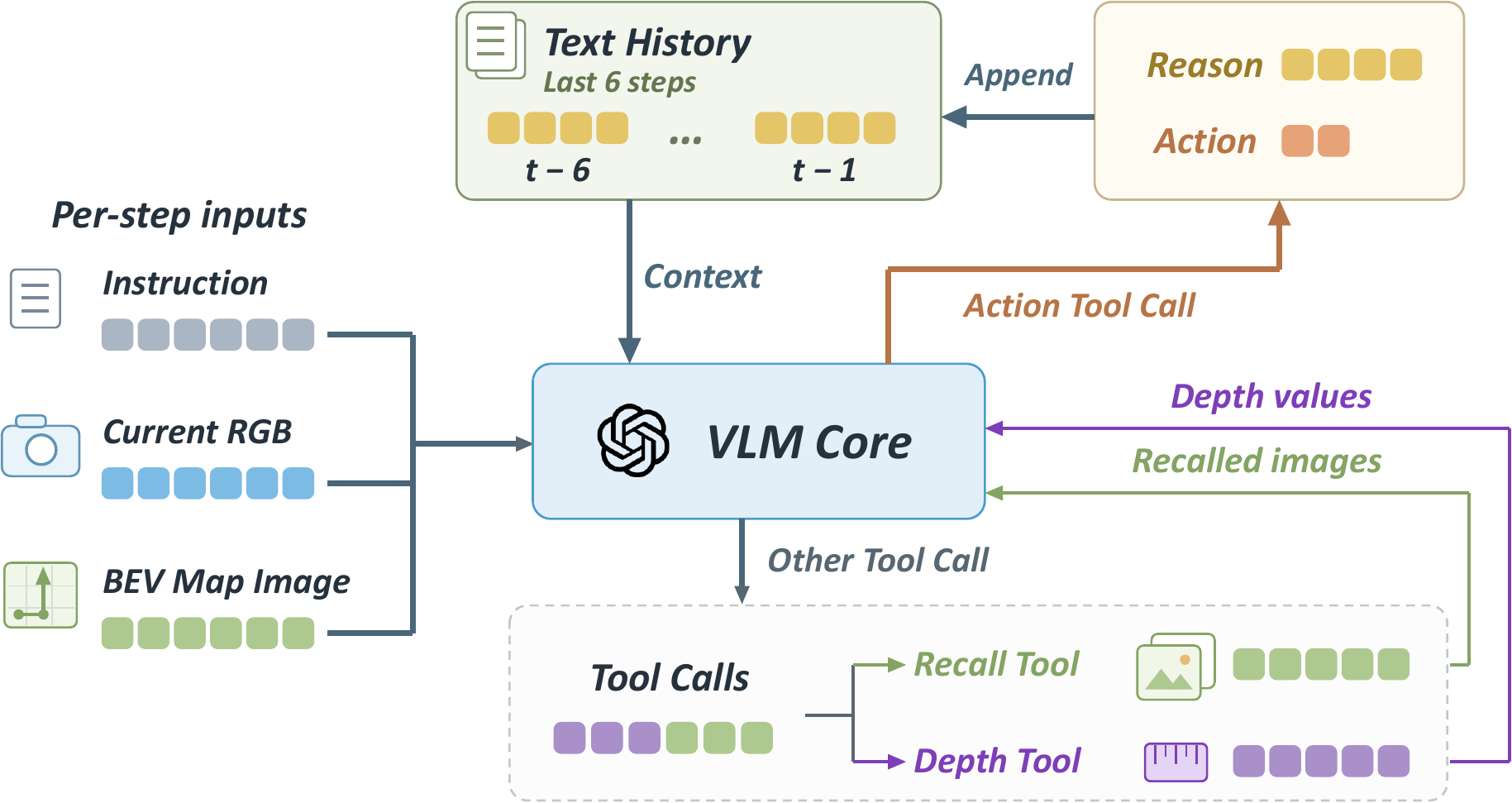}
\caption{Overview of AgenticNav's context management. At each step, the VLM receives the instruction, current RGB view, BEV map, and textual reasoning history, while the Recall and Depth Tools return visual or numerical evidence on demand.}
\label{fig:context_management}
\end{figure}

AgenticNav maintains within-episode \textbf{Agentic Memory}, $M_t=(H_t,B_t,C_t)$, comprising reasoning text history $H_t$, a compact BEV map image $B_t$, and a bounded image cache $C_t$ accessed through the Recall Tool. Fig.~\ref{fig:context_management} illustrates how these components support successive decisions.

\textbf{Reasoning text history.} Each completed action contributes a compact record containing its supplied \texttt{reason} and the corresponding action information. The prompt retains records from the last six steps:
\begin{equation}
H_t=\{(r_j,a_j)\}_{j=\max(0,t-6)}^{t-1}.
\end{equation}
Here $r_j$ is the recorded textual reason, and $a_j$ contains the selected target and executed motion. This preserves the connection between previous decisions and their rationales while keeping the history bounded.

\textbf{BEV map image.} $B_t$ summarizes the navigation trajectory and historical decision points, providing spatial context alongside the textual history. It also includes obstacle and traversability information from the underlying SLAM module. Its node indices also identify observations available for visual recall, as illustrated in Fig.~\ref{fig:method}.

\textbf{Selective visual recall tool.} The image cache $C_t$ stores past RGB observations. When visual detail from a previous decision point is needed, the VLM calls \texttt{recall}$(p,k)$, where $p$ identifies the decision point and $k$ selects a view:
\begin{equation}
R_t(p,k)=C_t[p,k].
\end{equation}
The requested image is appended to the current prompt. At each new step, the harness reconstructs the context from the instruction, current RGB views, $B_t$, and $H_t$; historical images are added on demand through subsequent recall calls.

\section{Experiments}
\label{sec:result}

\subsection{Simulation Experiments}

\paragraph{Setup and metrics.}
Simulation is conducted in the Habitat simulator. Following recent zero-shot VLN-CE evaluation protocols~\citep{qiao2025opennav,shi2025smartway,dai2026evonav}, we evaluate on the exact same 100 R2R-CE \texttt{val\_unseen} episodes. Following the same protocol, we report success rate (SR), oracle success rate (OSR), success weighted by path length (SPL), normalized Dynamic Time Warping (nDTW), and navigation error in meters (NE).

\paragraph{Baselines and results.}
Table~\ref{tab:main_results} compares AgenticNav with supervised and zero-shot VLN-CE methods, with $\ast$ marking our same-backbone reproductions for fair comparison. All reproduced runs keep the exact same experiment settings and VLM core as our method. Results show that AgenticNav-Gemini-3.7-Flash achieves the best zero-shot results across all five metrics, improving over SmartWay-Gemini-3.7-Flash from 54\% to 76\% SR and from 47.95\% to 66.50\% SPL. The gain suggests that the tool interface improves not only whether the agent eventually reaches the goal but also how efficiently it moves through the environment. Notably, AgenticNav also outperforms leading training-based methods such as LightNav-0, demonstrating that zero-shot approaches can match or even surpass supervised methods when equipped with a well-designed agent harness. With both Gemini-2.5-Pro and GPT-5.5, AgenticNav also surpasses the previous zero-shot state of the art, indicating that its gains are not tied to a single VLM backbone.
\begin{table}[!t]
\caption{Simulation comparison on R2R-CE.}
\label{tab:main_results}
\centering
\small
\setlength{\tabcolsep}{2.0pt}
\renewcommand{\arraystretch}{1.15}
\begin{tabularx}{\columnwidth}{@{}>{\raggedright\arraybackslash}X|| c c c c c@{}}
\toprule
\textbf{Method} & \textbf{SR$\uparrow$} & \textbf{SPL$\uparrow$} & \textbf{OSR$\uparrow$} & \textbf{nDTW$\uparrow$} & \textbf{NE (m)$\downarrow$} \\
\midrule
\rowcolor{rowgray}\multicolumn{6}{l}{\textit{VLN-CE with supervised learning}} \\
\hspace*{1.5em}CMA~\citep{hong2022bridging}       & 37 & 32.17 & 45 & 50.77 & 6.92 \\
\hspace*{1.5em}StreamVLN$^\ast$~\citep{wei2025streamvln} & 47 & 41.31 & 58 & 58.86 & 5.72 \\
\hspace*{1.5em}RecBERT~\citep{hong2022bridging}   & 48 & 43.22 & 57 & 54.81 & 5.8 \\
\hspace*{1.5em}NaVILA$^\ast$~\citep{cheng2024navila} & 49 & 44.22 & 60 & 59.83 & 6.36 \\
\hspace*{1.5em}ETPNav~\citep{an2023etpnav}        & 58 & 52.19 & 63 & 61.16 & 5.16 \\
\hspace*{1.5em}BEVBert~\citep{an2023bevbert}      & 60 & 53.41 & 64 & 61.40 & 5.13 \\
\hspace*{1.5em}AwareVLN$^\ast$~\citep{guo2026awarevln} & 63 & 56.34 & \textbf{79} & 65.11 & 4.37 \\
\hspace*{1.5em}LightNav-0$^\ast$~\citep{wang2026lightnav} & \textbf{67} & \textbf{62.71} & 72 & \textbf{66.80} & \textbf{4.03} \\
\midrule
\rowcolor{rowgray}\multicolumn{6}{l}{\textit{Zero-shot VLN-CE}} \\
\hspace*{1.5em}Random                          & 2 & 1.50 & 12 & 34.08 & 8.63 \\
\hspace*{1.5em}LXMERT~\citep{hong2022bridging} & 2 & 1.87 & 22 & 18.73 & 10.48 \\
\midrule
\multicolumn{6}{l}{\hspace*{0.75em}\textit{VLM backbone: GPT-5.5}} \\
\hspace*{1.5em}SmartWay$^\ast$~\citep{shi2025smartway} & 44 & 35.04 & 60 & 58.64 & \textbf{5.16} \\
\hspace*{1.5em}\textbf{AgenticNav (ours)} & \textbf{55} & \textbf{48.41} & \textbf{65} & \textbf{63.41} & 5.19 \\
\midrule
\multicolumn{6}{l}{\hspace*{0.75em}\textit{VLM backbone: Gemini-2.5-Pro}} \\
\hspace*{1.5em}Open-Nav$^\ast$~\citep{qiao2025opennav} & 27 & 23.22 & 32 & 54.51 & 6.46 \\
\hspace*{1.5em}EvoNav$^\ddagger$~\citep{dai2026evonav} & 43 & 37.77 & 51 & 62.38 & 5.04 \\
\hspace*{1.5em}SmartWay$^\ast$~\citep{shi2025smartway} & 52 & 41.70 & 62 & 62.10 & 5.28 \\
\hspace*{1.5em}\textbf{AgenticNav (ours)} & \textbf{58} & \textbf{42.91} & \textbf{76} & \textbf{63.49} & \textbf{5.00} \\
\midrule
\multicolumn{6}{l}{\hspace*{0.75em}\textit{VLM backbone: Gemini-3.7-Flash}} \\
\hspace*{1.5em}CA-Nav$^\ast$~\citep{chen2025canav} & 23 & 8.46 & 47 & 13.84 & 7.82 \\
\hspace*{1.5em}HSGM$^\ast$~\citep{li2026hsgm} & 26 & 21.47 & 31 & 54.51 & 6.16 \\
\hspace*{1.5em}Open-Nav$^\ast$~\citep{qiao2025opennav} & 31 & 24.93 & 40 & 52.02 & 6.41 \\
\hspace*{1.5em}LaViRA$^\ast$~\citep{ding2025lavira} & 38 & 31.71 & 49 & 55.69 & 5.52 \\
\hspace*{1.5em}SmartWay$^\ast$~\citep{shi2025smartway} & 54 & 47.95 & 60 & 65.93 & 4.68 \\
\hspace*{1.5em}\textbf{AgenticNav (ours)} & \textbf{76} & \textbf{66.50} & \textbf{78} & \textbf{69.38} & \textbf{3.48} \\
\bottomrule
\end{tabularx}
\vspace{-0.35em}
\parbox{\columnwidth}{\scriptsize $\ast$: Reproduced results; $\ddagger$: EvoNav, not episodic zero-shot because it retrieves previous-episode experience. Best results are in \textbf{bold}.}
\end{table}

\begin{table}[!t]
\caption{VLM-backbone ablation on R2R-CE. }
\label{tab:vlm_backbone}
\centering
\small
\setlength{\tabcolsep}{2pt}
\renewcommand{\arraystretch}{1.15}
\begin{tabularx}{\columnwidth}{@{}>{\raggedright\arraybackslash}X|| c c c c c@{}}
\toprule
\textbf{VLM Backbone} & \textbf{SR$\uparrow$} & \textbf{SPL$\uparrow$} & \textbf{OSR$\uparrow$} & \textbf{nDTW$\uparrow$} & \textbf{NE (m)$\downarrow$} \\
\midrule
GPT-5.5                         & 55 & 48.41 & 65 & 63.41 & 5.19 \\
Gemini-2.5-Pro & 58 & 42.91 & 76 & 63.49 & 5.00 \\
Gemini-3.7-Flash & \textbf{76} & \textbf{66.50} & \textbf{78} & \textbf{69.38} & \textbf{3.48} \\
\resizebox{\linewidth}{!}{DeepSeek-V4-Flash Vision-Exp} & 49 & 36.48 & 64 & 61.18 & 5.67 \\
Qwen3.8-27B & 48 & 33.57 & 72 & 61.58 & 5.65 \\
\bottomrule
\end{tabularx}
\end{table}

\begin{figure}[!t]
\centering
\includegraphics[width=\columnwidth]{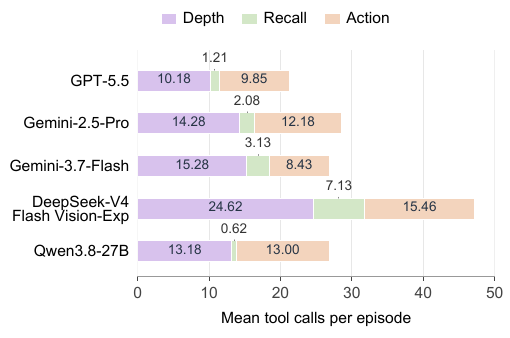}
\caption{Mean calls to the Depth, Recall, and Action Tools per episode across VLM backbones, averaged over 100 navigation episodes per backbone. Stop calls are excluded.}
\label{fig:vlm_tool_calls}
\end{figure}

\paragraph{VLM cores.}
Table~\ref{tab:vlm_backbone} evaluates VLM backbones with AgenticNav. Gemini-3.7-Flash is strongest overall, reaching 76\% SR and 66.50\% SPL, while Gemini-2.5-Pro, GPT-5.5, DeepSeek, and Qwen3.8-27B retain substantial zero-shot navigation capability. Fig.~\ref{fig:vlm_tool_calls} reports the average number of tool calls per episode for different VLM backbones. The backbones differ in both their total number of calls and their preferences for individual tools, while all three tools are actively used across backbones. The pattern suggests that better VLM reasoning improves final stopping and path efficiency, but the tool structure itself remains useful even with small backbones like Qwen3.8-27B.

\begin{figure*}[!t]
\begin{minipage}[t]{\dimexpr(\textwidth-\columnsep)/2\relax}
\vspace{0pt}
\centering
\includegraphics[width=\columnwidth]{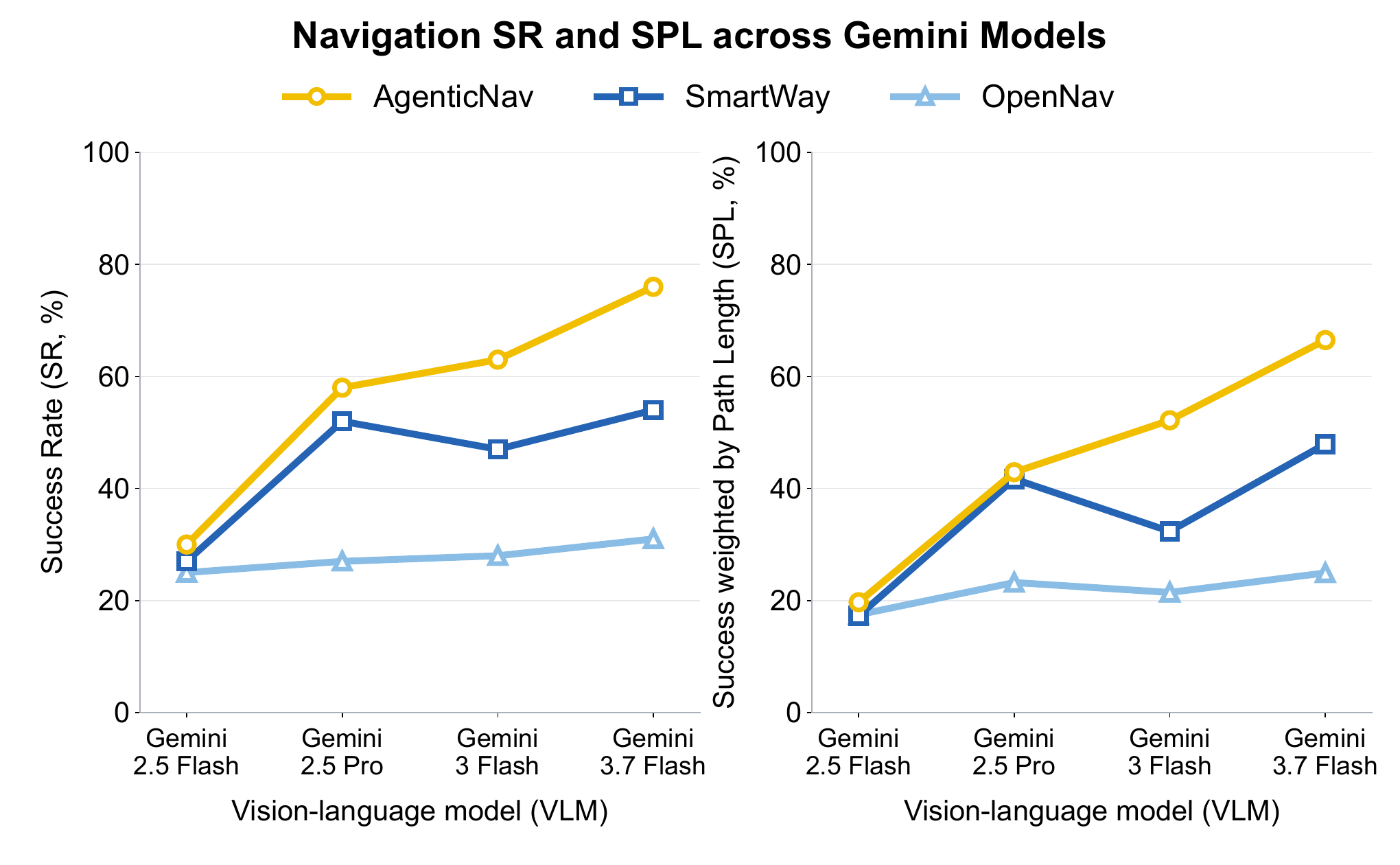}
\caption{Navigation performance across Gemini backbones on R2R-CE. AgenticNav exhibits sustained gains in both success rate and path efficiency, while the compared methods show more limited gains beyond Gemini-2.5-Pro.}
\label{fig:backbone_scaling}
\end{minipage}\hfill
\begin{minipage}[t]{\dimexpr(\textwidth-\columnsep)/2\relax}
\vspace{0pt}
\centering
\includegraphics[width=\columnwidth]{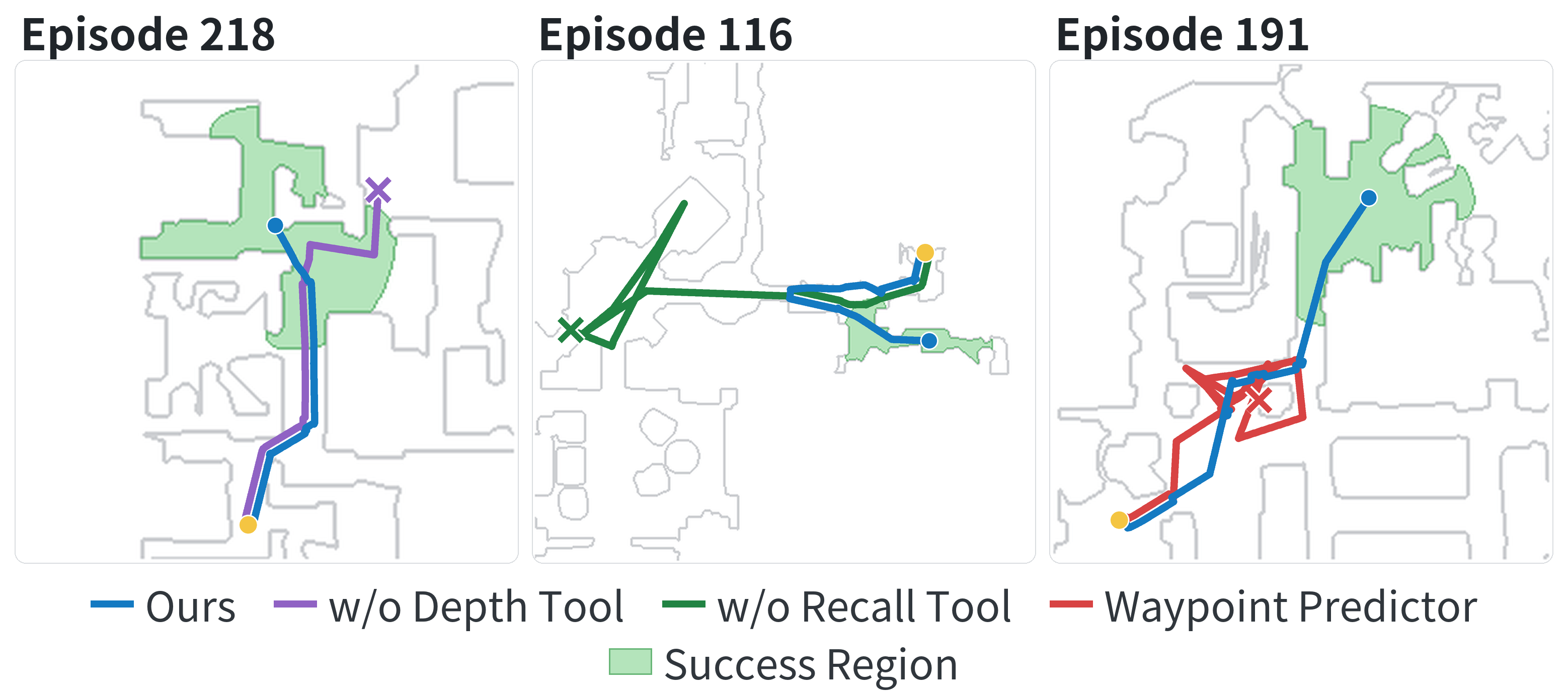}
\caption{Qualitative case analysis for tool component ablations on RCR-CE. In Episode 218, the variant without Depth Tool struggles to estimate distances accurately and mistakenly stop outside the 3\,m goal region, illustrating the role of the Depth Tool in accurate stopping. In Episode 116, Recall Tool helps the agent correct its mistake and backtrack after passing the goal region. In Episode 191, the variant using the Waypoint Predictor gets stuck midway due to unsuitable waypoints, while methods using the Action Tool successfully navigate the complex room layout.}
\label{fig:tool_ablation_routes}
\end{minipage}
\end{figure*}

\paragraph{Scaling with VLM Backbones}
Fig.~\ref{fig:backbone_scaling} shows sustained gains in success and path efficiency for AgenticNav across the evaluated Gemini backbones, whereas SmartWay and Open-Nav obtain limited improvements beyond Gemini-2.5-Pro. From Gemini-2.5-Pro to Gemini-3.7-Flash, AgenticNav gains 18.00/23.59 percentage points in SR/SPL, compared with 2.00/6.25 for SmartWay and 4.00/1.71 for Open-Nav. These results suggest that AgenticNav translates backbone improvements into navigation gains more effectively than the compared methods, which is a key design objective for an agent harness. It also demonstrates AgenticNav's potential to scale with future advances in VLMs and translate their evolving capabilities into further navigation gains.

\paragraph{Memory and Context Management}
Table~\ref{tab:context_management} examines how VLM memory and context are organized while keeping the remaining navigation settings fixed. Removing action reasoning text reduces SR by 13 points. Replacing action reasoning with a structured task-progress record specifying completed instruction clauses, the current subgoal, and the stopping condition yields little recovery. We attribute this difference to the way errors are carried forward: a reasoning history is less likely to propagate wrong judgments, whereas an explicit task-progress record can carry a mistaken assessment forward and repeatedly influence subsequent decisions. Without the BEV summary, SR decreases from 76\% to 70\%. Accumulating every historical image also lowers SR/SPL by 10/12.45 points. These results suggest that a compact visual context combining the BEV map with selective recall is more effective than accumulating a large history of images.

\begin{table}[!t]
\caption{Memory and Context-management ablation on R2R-CE}
\label{tab:context_management}
\centering
\small
\setlength{\tabcolsep}{4pt}
\renewcommand{\arraystretch}{1.12}
\begin{tabularx}{\columnwidth}{@{}>{\raggedright\arraybackslash}X c c@{}}
\toprule
\textbf{Context Variant} & \textbf{SR$\uparrow$} & \textbf{SPL$\uparrow$} \\
\midrule
\textbf{AgenticNav-Gemini-3.7-Flash (Full) }& \textbf{76} & \textbf{66.50} \\
\quad w/o action reasoning history text& 63 & 48.49 \\
\quad replace action reasoning with task progress text & 64 & 52.56 \\
\quad w/o BEV map image& 70 & 59.46 \\
\quad accumulate all historical images & 66 & 54.05 \\

\bottomrule
\end{tabularx}
\end{table}

\paragraph{Tool components.}
Table~\ref{tab:ablation} compares variants with individual tools removed or replaced. Replacing the Action Tool with a Waypoint Predictor lowers navigation performance, supporting the benefit of our tool design. Replacing the Depth Tool with Depth Image Input also reduces performance, suggesting that querying depth at selected locations is more effective than asking the VLM to interpret a full depth image. Removing the Depth Tool or the Recall Tool further reduces performance, highlighting the importance of distance information and access to past observations. Fig.~\ref{fig:tool_ablation_routes} illustrates how these tools support accurate stopping, backtracking, and target selection in three navigation episodes.

\subsection{Real-world Experiments}

\paragraph{Setup and baseline.}
We deploy AgenticNav on two robotic platforms: a Unitree Go2W quadruped robot and an omnidirectional four-wheel robot. Both platforms are equipped with four Intel RealSense D435 RGB-D cameras and LiDAR, and use FAST-LIO for localization. To generate local trajectories toward the targets provided by the Action Tool, we use TEB~\citep{rosmann2017integrated} on the four-wheel robot and SCAN-Planner~\citep{zheng2026scan} on the Unitree Go2W. We define 35 real-world episodes spanning laboratory, office, and outdoor plaza scenes, covering sign understanding, long-range navigation, low-light nighttime conditions, etc. Compared with simulation, these environments are more diverse and better reflect practical navigation conditions. We compare against SmartWay, the strongest reproducible baseline in our simulation experiments. AgenticNav uses one or four camera views, while SmartWay collects 12 views by rotating in place to meet the input requirements of its waypoint predictor. For each episode, all methods use the same robot platform, route, instruction, and evaluation settings.

\begin{table*}[!t]
\caption{AgenticNav tool component ablations on R2R-CE. }
\label{tab:ablation}
\centering
\small
\setlength{\tabcolsep}{2pt}
\renewcommand{\arraystretch}{1.15}
\begin{tabularx}{\textwidth}{@{}>{\raggedright\arraybackslash}X|| c c c c c | c c c c c@{}}
\toprule
\textbf{Variant} & \shortstack{\textbf{Action}\\\textbf{Tool}} & \shortstack{\textbf{Depth}\\\textbf{Tool}} & \shortstack{\textbf{Recall}\\\textbf{Tool}} & \shortstack{\textbf{Depth}\\\textbf{Image}} & \shortstack{\textbf{Waypoint}\\\textbf{Predictor}} & \textbf{SR$\uparrow$} & \textbf{SPL$\uparrow$} & \textbf{OSR$\uparrow$} & \textbf{nDTW$\uparrow$} & \textbf{NE (m)$\downarrow$} \\
\midrule
\textbf{AgenticNav-Gemini-3.7-Flash (Full) }& \cmark & \cmark & \cmark & \xmark & \xmark & \textbf{76} & \textbf{66.50} & \textbf{78} & 69.38 & \textbf{3.48} \\
\quad replace Action Tool with Waypoint Predictor & \xmark & \cmark & \cmark & \xmark & \cmark & 66 & 57.73 & 75 & \textbf{69.82} & 3.91 \\
\quad w/o Depth Tool & \cmark & \xmark & \cmark & \xmark & \xmark & 60 & 47.14 & 76 & 65.53 & 5.03 \\
\quad replace Depth Tool with Depth Image Input & \cmark & \xmark & \cmark & \cmark & \xmark & 69 & 55.79 & 76 & 65.68 & 4.42 \\
\quad w/o Recall Tool & \cmark & \cmark & \xmark & \xmark & \xmark & 64 & 47.47 & 77 & 64.20 & 5.10 \\
\bottomrule
\end{tabularx}
\end{table*}

\begin{table*}[!t]
\caption{Real-world comparison. }
\label{tab:real_world}
\centering
\small
\setlength{\tabcolsep}{3.8pt}
\renewcommand{\arraystretch}{1.15}
\begin{tabularx}{\textwidth}{l||Y Y|Y Y|Y Y|Y Y}
\toprule
\multirow{2}{*}{\textbf{Method}} &
\multicolumn{2}{c|}{Laboratory} &
\multicolumn{2}{c|}{Office} &
\multicolumn{2}{c|}{Plaza} &
\multicolumn{2}{c}{\textbf{Overall}} \\
\cmidrule(lr){2-3}\cmidrule(lr){4-5}\cmidrule(lr){6-7}\cmidrule(lr){8-9}
& \textbf{SR$\uparrow$} & \textbf{NE (m)$\downarrow$}
& \textbf{SR$\uparrow$} & \textbf{NE (m)$\downarrow$}
& \textbf{SR$\uparrow$} & \textbf{NE (m)$\downarrow$}
& \textbf{SR$\uparrow$} & \textbf{NE (m)$\downarrow$} \\
\midrule
SmartWay-GPT-5.5 (12 view)~\citep{shi2025smartway} & 44.4 & 2.60 & 17.6 & 2.62 & 11.1 & 6.98 & 22.9 & 3.74 \\
\textbf{AgenticNav-GPT-5.5 (1 view) (ours)} & 55.6 & 2.41 & 35.3 & 2.18 & 22.2 & 5.75 & 37.1 & 3.16 \\
\textbf{AgenticNav-GPT-5.5 (4 view) (ours)}
                                  & \textbf{66.7} & \textbf{1.96}
                                  & \textbf{47.1} & \textbf{1.85}
                                  & \textbf{55.6} & \textbf{4.29}
                                  & \textbf{54.3} & \textbf{2.51} \\
\bottomrule
\end{tabularx}
\end{table*}

\begin{figure*}[!t]
\centering
\includegraphics[width=\linewidth]{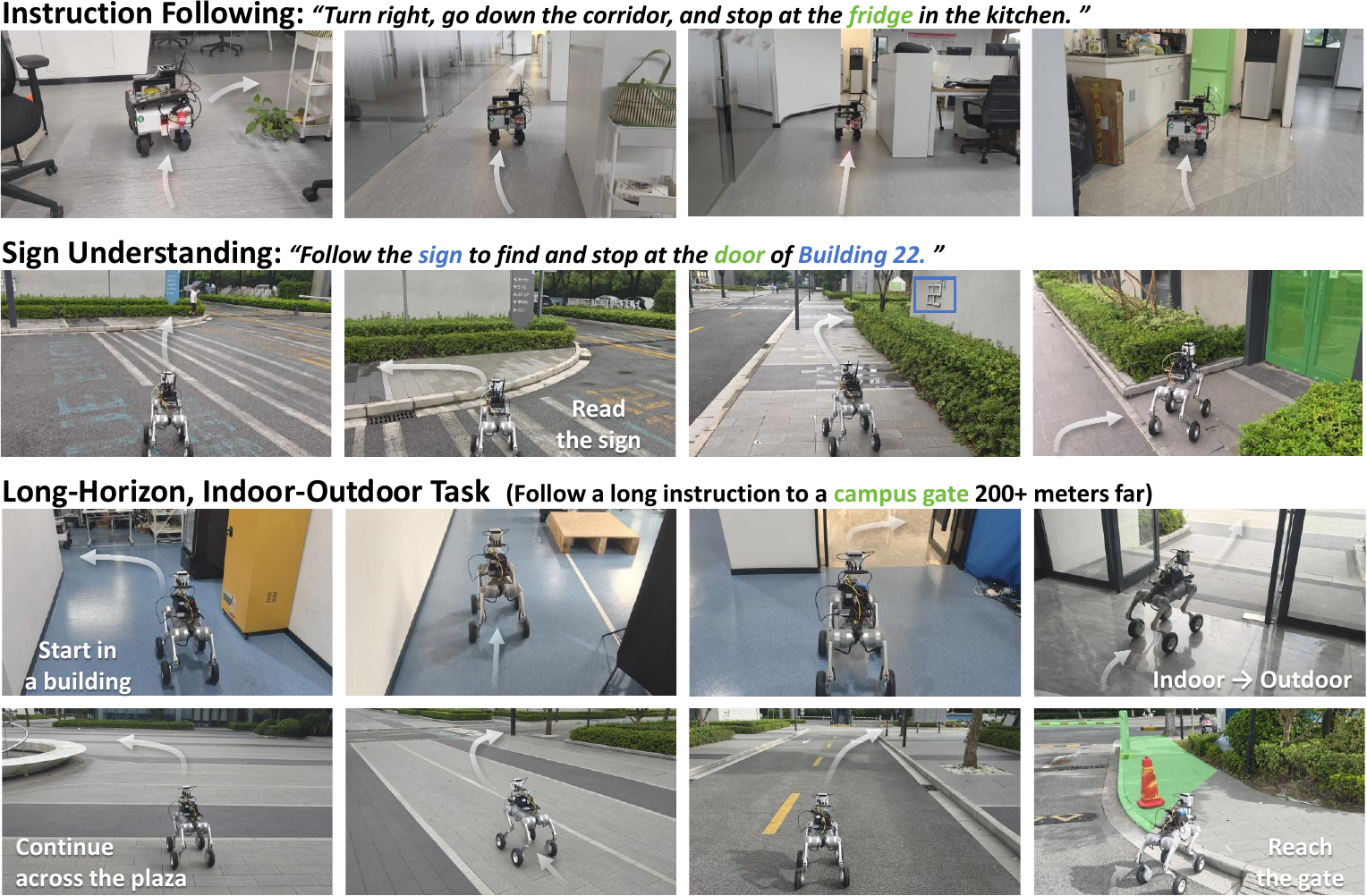}
\caption{Real-world navigation demonstrations. Details can be found in the supplementary video.}
\label{fig:real_tool_demos}
\end{figure*}

\paragraph{Quantitative results.}
Table~\ref{tab:real_world} shows that AgenticNav outperforms SmartWay in both one-view and four-view settings. One view improves overall SR/NE from 22.9\%/3.74m to 37.1\%/3.16m despite using fewer observations than SmartWay. Four views reach 54.3\%/2.51m, more than doubling SmartWay's success rate and giving the largest gain in the outdoor yard, where wide turns and distant goals are common. Extra views benefit our method by exposing more semantic pixel targets through the Action Tool, allowing the robot to execute larger turns. SmartWay often fails when its learned waypoint predictor misses the instruction-relevant direction or cannot precisely select the target point, reflecting the generalization limitation of trained waypoint modules and highlighting the advantage of our purely zero-shot approach.

\paragraph{Qualitative analysis.}
Fig.~\ref{fig:real_tool_demos} presents three real-world navigation tasks. In the top row, the robot follows the instruction through a corridor to the fridge in the kitchen. In the middle row, it reads the sign and makes several turns to reach the door of Building 22. In the bottom two rows, it completes a long-horizon indoor-to-outdoor task over more than 200\,m, leaving the building, crossing the plaza, and reaching the campus gate. These cases demonstrate the adaptability and practical utility of AgenticNav in real-world environments, as well as its ability to generalize across different robotic platforms and environments.


\section{Failure Analysis}
\label{sec:limitations}
\begin{figure}[!t]
\centering
\includegraphics[width=\columnwidth]{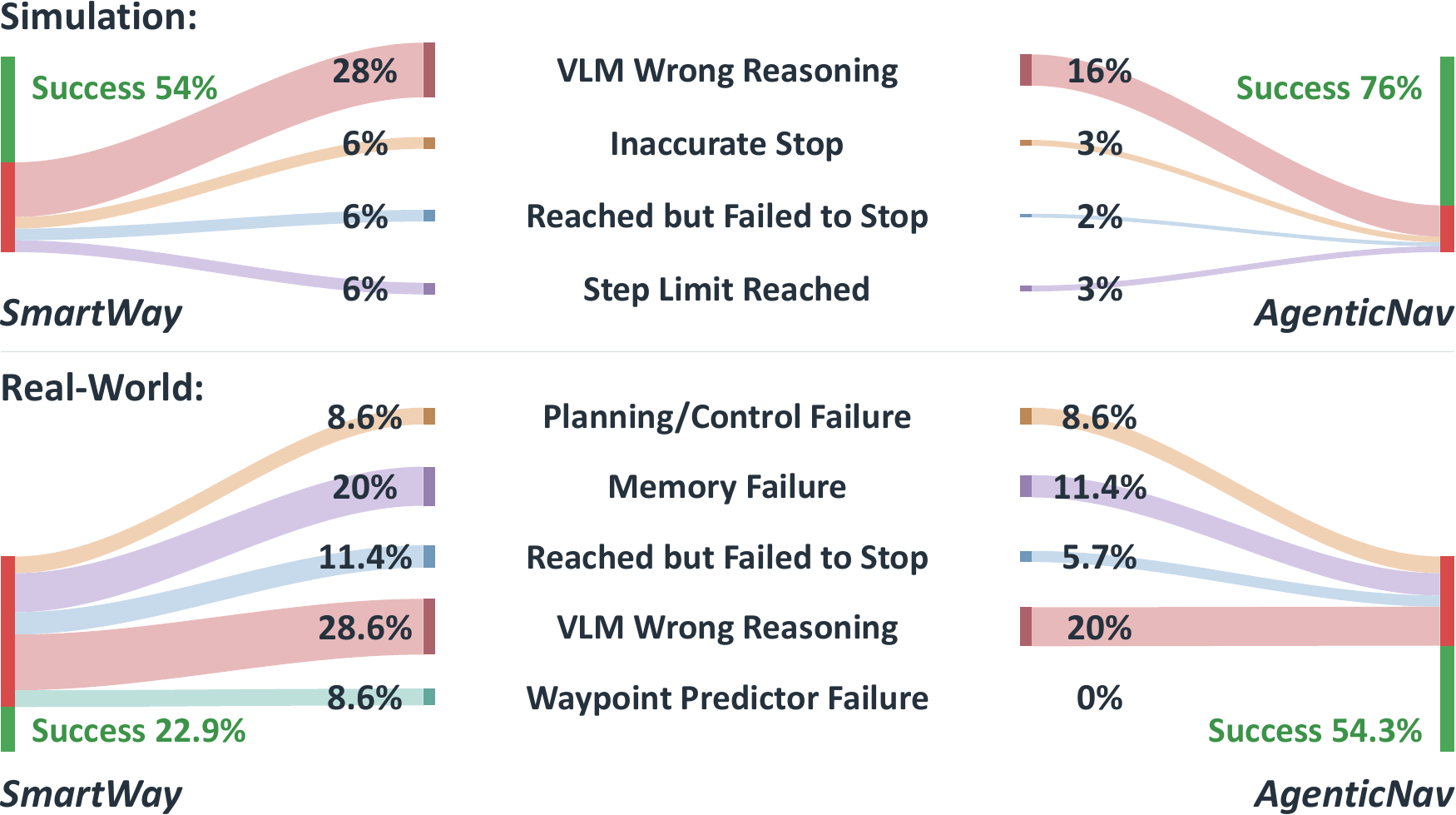}
\caption{Failure analysis in simulation (top) and real-world experiments (bottom). Percentages are over all trials per method.}
\label{fig:failure_analysis}
\end{figure}

Fig.~\ref{fig:failure_analysis} provides a failure breakdown comparing AgenticNav and Smartway in simulation and real-world experiments. Overall, AgenticNav suppresses VLM reasoning errors, demonstrating that our proposed agentic harness more effectively unlocks the potential of VLMs. Stop-related failures are reduced by more than half compared to SmartWay, underscoring the advantage of exact spatial distance provided by depth tool. In real-world experiments, memory-related failures drop nearly by half, validating the effectiveness of agentic memory in long-horizon navigation tasks. Furthermore, while the waypoint predictor produces unreliable predictions in real world experiments, our action tool completely eliminates waypoint-predictor-induced failures. Overall, these improvements confirm that AgenticNav effectively alleviates interface-level constraints, leaving core VLM reasoning as the primary residual error source.

However, AgenticNav still remains bounded by the VLM's decision quality. Since our experiments show that AgenticNav achieves consistent gains in navigation performance as VLMs improve, we may look forward to further advances in foundation models bringing continued improvements to AgenticNav.

\section{Conclusion}
\label{sec:conclusion}
We presented AgenticNav, a zero-shot VLN-CE harness that treats navigation as tool calling between a VLM and the environment. Instead of relying on a trained waypoint predictor or accumulating long prompt histories, AgenticNav exposes three grounded capabilities to the model: pixel-level action selection, targeted metric depth queries, and selective visual memory recall. Simulation, real-world, and ablation experiments collectively demonstrate the effectiveness and advantages of our method. These results suggest that, for foundation-model navigation, the design of the action, perception, and memory harness is as important as the choice of the VLM itself. Our results also suggest that a navigation harness should be designed not only for current VLM capabilities, but also to translate future backbone improvements into better embodied performance.


\bibliographystyle{IEEEtran-short}
\bibliography{references}  

\end{document}